\documentclass[10pt,twocolumn,letterpaper]{article}

\usepackage{amsmath,amsfonts,bm}

\def\eqref#1{equation~\ref{#1}}

\def\1{\bm{1}}

\DeclareMathAlphabet{\mathsfit}{\encodingdefault}{\sfdefault}{m}{sl}
\SetMathAlphabet{\mathsfit}{bold}{\encodingdefault}{\sfdefault}{bx}{n}

\usepackage{cvpr}
\usepackage{times}
\usepackage{epsfig}
\usepackage{graphicx}
\usepackage{amsmath}
\usepackage{amssymb}

\usepackage{authblk}

\usepackage[algo2e,ruled,vlined,linesnumbered]{algorithm2e}
\usepackage{url} 
\usepackage[export]{adjustbox}
\usepackage{booktabs}
\usepackage[font=normalsize]{subfig}

\usepackage{url}
\usepackage{xcolor}

\usepackage[pagebackref=true,breaklinks=true,letterpaper=true,colorlinks,bookmarks=false]{hyperref}

\cvprfinalcopy 

\def\cvprPaperID{301} 

\ifcvprfinal\fi
\begin{document}

\newcommand{\HCOT}{HCOT\xspace}
\newcommand{\COT}{COT\xspace}
\newcommand{\HCE}{HCE\xspace}
\newcommand{\hide}[1]{}
\newcommand{\comment}[1]{{\color{red}#1}}

\title{Learning with Hierarchical Complement Objective}



\author[1]{Hao-Yun Chen}
\author[1]{Li-Huang Tsai}
\author[1, 2]{Shih-Chieh Chang}
\author[3]{Jia-Yu Pan}
\author[3]{Yu-Ting Chen}
\author[3]{Wei Wei}
\author[3]{Da-Cheng Juan}

\affil[1]{\footnotesize Department of Computer Science, National Tsing-Hua University, Hsinchu, Taiwan}
\affil[2]{\footnotesize Electronic and Optoelectronic System Research Laboratories, ITRI, Hsinchu, Taiwan}
\affil[3]{\footnotesize Google Research, Mountain View, CA, USA}


\maketitle

\begin{abstract}
Label hierarchies widely exist in many vision-related problems, ranging from explicit label hierarchies existed in image classification to latent label hierarchies existed in semantic segmentation. Nevertheless, state-of-the-art methods often deploy cross-entropy loss that implicitly assumes class labels to be exclusive and thus independence from each other. Motivated by the fact that classes from the same parental category usually share certain similarity, we design a new training diagram called Hierarchical Complement Objective Training (\HCOT) that leverages the information from label hierarchy. \HCOT maximizes the probability of the ground truth class, and at the same time, neutralizes the probabilities of rest of the classes in a hierarchical fashion, making the model take advantage of the label hierarchy explicitly. The proposed \HCOT is evaluated on both image classification and semantic segmentation tasks. Experimental results confirm that \HCOT outperforms state-of-the-art models in CIFAR-100, ImageNet-2012, and PASCAL-Context. The study further demonstrates that \HCOT can be applied on tasks with latent label hierarchies, which is a common characteristic in many machine learning tasks.

\end{abstract}



\section{Introduction}
\label{sec:Introduction}

\begin{figure*}[!t]
\centering
\subfloat[Baseline (cross-entropy)]{
    \begin{minipage}[t]{0.33\linewidth}
        \centering
        \includegraphics[width=\linewidth]{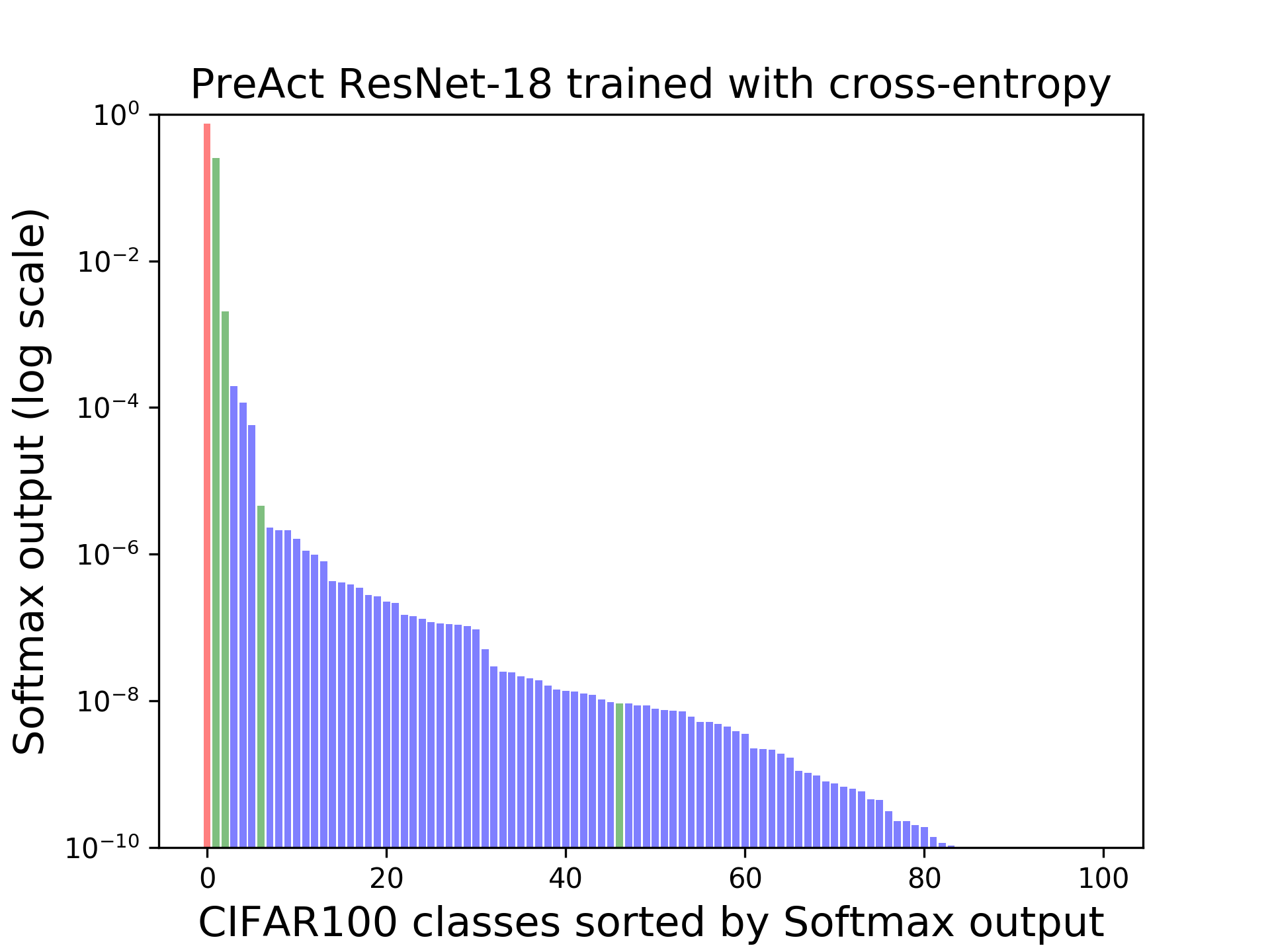}\\
        \vspace{0.02cm}
        \label{fig:probability_baseline}
    \end{minipage}
    }
\subfloat[COT]{
    \begin{minipage}[t]{0.33\linewidth}
        \centering
        \includegraphics[width=\linewidth]{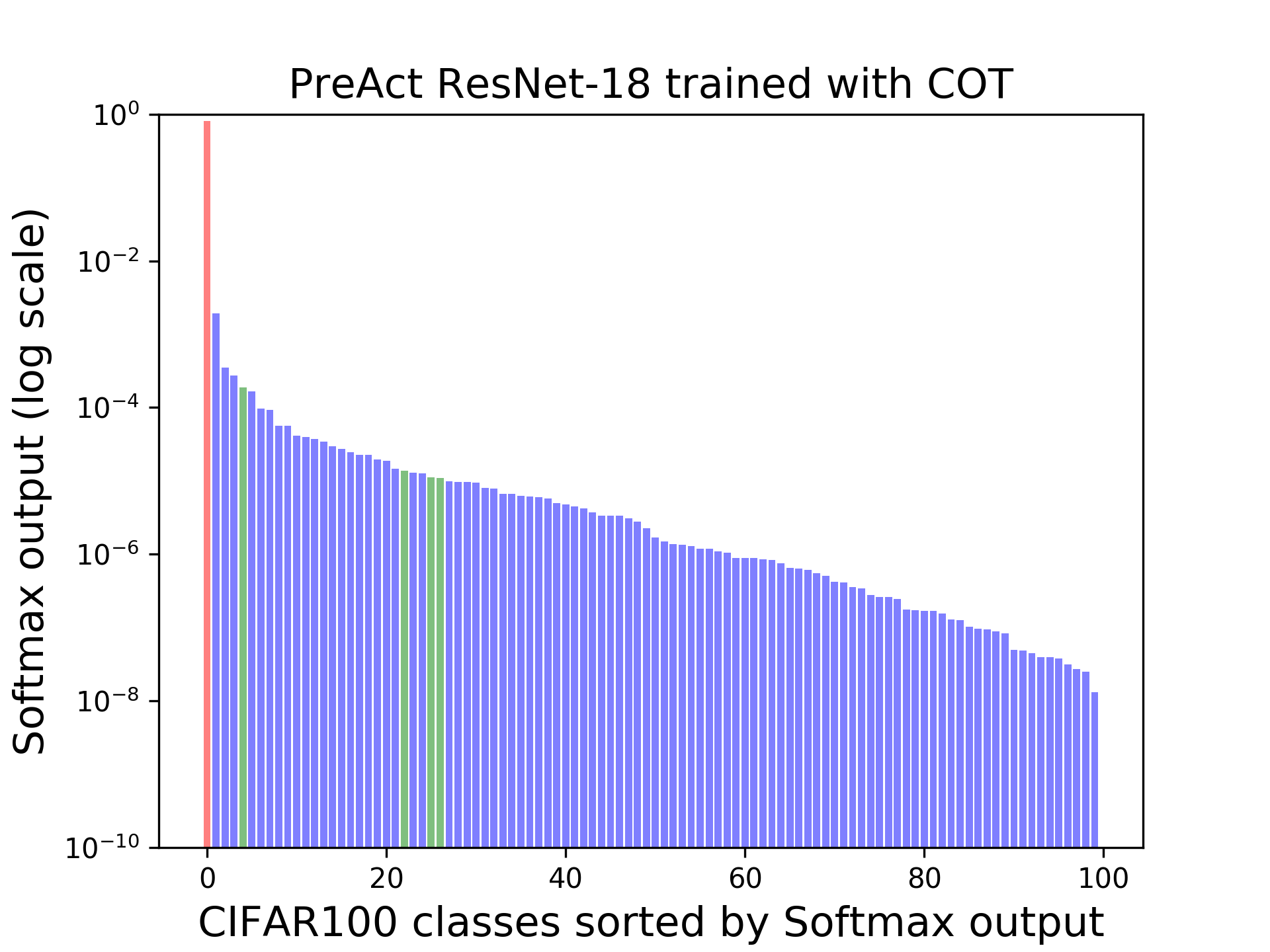}\\
        \vspace{0.02cm}
        \label{fig:probability_COT}
    \end{minipage}
    }
\subfloat[\HCOT]{
    \begin{minipage}[t]{0.33\linewidth}
        \centering
        \includegraphics[width=\linewidth]{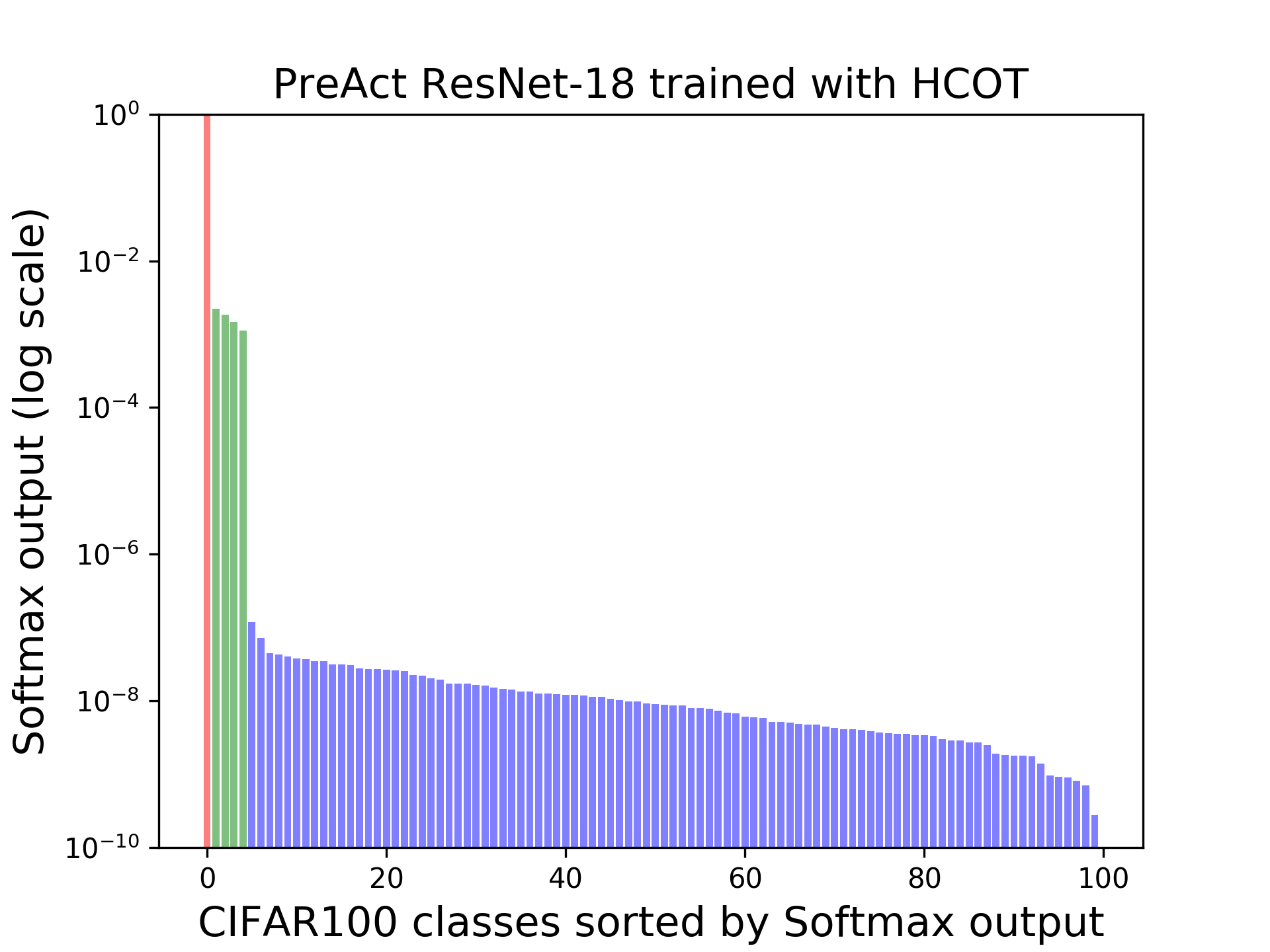}\\
        \vspace{0.02cm}
        \label{fig:probability_HCOT}
    \end{minipage}
}

\centering
\caption{Sorted predicted probabilities (denoted as $\hat{y}$) from three different training paradigms evaluated on CIFAR-100 dataset using PreAct ResNet-18. The red bar indicates the probability of the ground-truth (denoted as $\hat{y}_g$), the green bars are the probabilities of classes in the same parental category as the ground-truth (denoted as $\hat{y}_{G\setminus\{g\}}$), and blue bars are the probabilities of the rest classes (denoted as $\hat{y}_{K \setminus G}$, see Sec.~\ref{sec:Hierarchical Complement Entropy} for detailed notation definition). Notice the ``staircase shape'' in (c) showing the significant difference between $\hat{y}_g$ and $\hat{y}_{G\setminus\{g\}}$, and then between $\hat{y}_{G\setminus\{g\}}$ and $\hat{y}_{K \setminus G}$, which confirms \HCOT well captures the label hierarchy.}
\label{FIG:probability}
\end{figure*}


Many machine learning tasks involve making predictions on classes that have an inherent hierarchical structure. One example would be image classification with hierarchical categories, where a category shares the same \textit{parental category} with other ones. For example, the categories with label ``dog'' and ``cat'' might share a common parental category ``pet'', which forms a \textit{explicit label hierarchy}. Another example would be in the task of semantic segmentation, where ``beach'', and ``sea'' are under the same theme ``scenery'' which forms a \textit{latent label hierarchy}, while ``people'', and ``pets'' forms another one of ``portrait.'' In this work, we call a parental category a \textit{coarse(-level) category}, while a category under a coarse category is called a \textit{fine(-level) category}.


Many successful deep learning models are built and trained with cross-entropy loss that assumes
prediction classes to be mutually independent. This assumption works well for many tasks such
as traditional image classifications where no hierarchical information is present. In the explicitly hierarchical setting, however, one problem is that learning with objectives that pose such a strong assumption makes the model
difficult to utilize the hierarchical structure in the label space. Another challenge in modeling hierarchical labels is that many tasks sometime exhibit latent label hierarchy. Take semantic segmentation for example, an inherent hierarchical structure has been explored by~\cite{Zhang_2018_CVPR} as ``'global context''. However, the dataset itself does not contain hierarchical information. 




In this paper, we develop a technique that is capable of leveraging the information in a label hierarchy, through proposing a new training objective. Our proposed technique is different from previous methods~\cite{Guo2018, Murdock_2016_CVPR, Yan_2015_ICCV, Zhang_2018_CVPR} which exploit the label hierarchy by changing model architectures but not the objectives. The general idea we propose is to penalize incorrect classes at different granularity levels: the classes that are ``obviously wrong''---different from not only the ground truth but also the parental category of ground truth---should receive larger penalty than the ones that share the same parental categories of ground truth. Such a mechanism allows us to take advantage of the information in the label hierarchy during training.

To achieve this goal of training with hierarchy information, we introduce the concept of Complement Objective Training (COT)~\cite{Chen2019ImprovingAR, chen2018complement} into label hierarchy. In COT, the probability of the correct class is maximized by a primary objective (i.e., cross-entropy), while the probabilities of incorrect classes are neutralized by a complement objective~\cite{chen2018complement}. This training paradigm aims at widening the gaps between the predicted probability value of the ground truth and those of the incorrect classes. In this paper, we propose \textbf{H}ierarchical \textbf{C}omplement \textbf{O}bjective \textbf{T}raining (\HCOT) with a novel complement objective called ``Hierarchical Complement Entropy'' (defined in Sec.~\ref{sec:Hierarchical Complement Entropy}), by applying the idea of the complement objective on both the fine-level class and its corresponding coarse-level class. 

\HCOT learns the class probabilities by three folds: (a) maximizing the predicted probability of ground truth, (b) neutralizing the predicted probabilities of incorrect classes sharing the same coarse-level category as the ground truth, and (c) further penalizing others that are on different branches (in the label hierarchy) to the ground-truth class. Figure~\ref{FIG:probability} illustrates the general idea of \HCOT compared to cross-entropy and COT, which shows \HCOT leads to both confident prediction for the ground-truth class and the predicted distribution that better reflects the label hierarchy (and therefore closer to the true data distribution). Particularly, the probability mass of the classes belonging to the parental category of the ground truth (in green) to be significantly higher than the rest of the classes (in blue). In other words, the model is trained to strongly penalize the “obviously wrong” classes that are completely irrelevant to both the ground-truth class and other classes belonging to the same parental category.

We conduct \HCOT on two important problems: image classification and semantic segmentation. Experimental results show that models trained with the Hierarchical complement entropy achieve significantly better performance over both cross-entropy and COT, across a wide range of state-of-the-art methods. We also show that \HCOT improves model performance when predicting the coarse-level classes. And finally, we show that \HCOT can deal with not only tasks with \textit{explicit label hierarchy} but also those with \textit{latent label hierarchy}. To the best of our knowledge, \HCOT is the first paradigm that trains deep neural models using an objective to leverage information from a label hierarchy, and leads to significant performance improvement.

\section{Background}
\label{sec:RelatedWork}





\paragraph{Learning Label Hierarchy.} To leverage label hierarchies from data has been explored for general purposes for years.~\cite{Goo2016TaxonomyRegularizedSD} tried to exploit discriminative properties from label hierarchies by regularization layers.~\cite{Bilal2017} studied how label hierarchies can help deep neural networks by utilizing the confusion patterns of fine categories to follow a hierarchical structure over the classes. For fine-grained image classification,~\cite{7298880} augments the fine-grained data with auxiliary images labeled by coarse classes, which exploits a regularization between fine-grained recognition models and coarse-grained recognition models. However, the above-mentioned approaches are usually not compatible with the cutting-edge deep models~\cite{He_2016_CVPR, Hu_2018_CVPR} and data augmentations~\cite{cutout, mixup} proposed in recent years. State-of-the-art methods in hierarchical problems have since then tend to adopt methods that ignore
hierarchical information during training.



\paragraph{Explicit Label Hierarchy.} Many tasks exhibit explicit label hierarchy that are presented as part of the dataset. Explicit hierarchical structures exist among the class labels for a wide range of problems. Taking visual recognition as an example, there have been many prior arts on non-neural models focused on exploiting the hierarchical structure in categories~\cite{TOUSCH2012333}. For neural models, HD-CNN~\cite{Yan_2015_ICCV} is an early work using the category hierarchy to improve performance over the flat N-way deep-network classifiers. The network architecture of HD-CNN contains a coarse component and several fine-grained components for learning from labels of different levels. Unlike HD-CNN which uses one fixed model, Blockout~\cite{Murdock_2016_CVPR} uses a regularization framework that learns both the model parameters and the sub-networks within a deep neural network, to capture the information in a label hierarchy. Another prior art named CNN-RNN~\cite{Guo2018} combines the CNN-based classifier with a Recurrent Neural Network to exploit the hierarchical relationship, sequentially from the coarse categories to the fine ones. All of the above-mentioned approaches rely on modifying model architectures to capture the hierarchical structures among the class labels. This raises an intriguing question:  Is it possible to design a training objective, rather than proposing a new model architecture, for a deep neural network to effectively capture the information contained in a label hierarchy?

\paragraph{Latent Label Hierarchy.} Another group of tasks are rather exclusive on the hierarchical information but has an underlying assumption on an inherent label structure. Semantic segmentation is one of such tasks where co-occurrence of the class labels forms a latent label hierarchy. This hierarchy is not directly observed in the data but can be inferred
from the data. In semantic segmentation, the goal is to assign a semantic label to each pixel of an image. Typically, when training a deep network model for semantic segmentation, the information of individual pixels are usually taken in isolation.  That is, the per-pixel cross-entropy loss is calculated for an image, with respect to the ground truth labels. To consider the global information, EncNet~\cite{Zhang_2018_CVPR} first utilizes the semantic context of scenes by exploiting model structures and provides a strong baseline in semantic segmentation. However, we argue that the potential of leveraging global information on the labeling space is still not discovered.

\section{Hierarchical Complement Objective Training}
\label{sec:Hierarchical Complement Entropy}
In this section, we introduce the proposed Hierarchical Complement Objective Training (\HCOT), which is a new training paradigm for leveraging information in a label hierarchy. Specifically, a novel training objective, Hierarchical Complement Entropy (HCE), is defined as the complement objective for \HCOT. In the following, we first review the concept of the complement objective, and then provide the mathematical formulation of HCE.  


\paragraph{Complement Objective.}
In Complement Objective Training (COT)~\cite{chen2018complement}, a neural model is trained with both a primary objective and a complement objective: the primary objective (e.g., cross-entropy) is for maximizing the predicted probability of the ground-truth class, whereas the complement objective (e.g., complement entropy~\cite{chen2018complement}) is designed to neutralize the predicted probabilities of incorrect classes, which intuitively makes a model more confident about the ground-truth class. Eq(\ref{eq:CE}) gives the definition of the complement entropy:

\begin{equation}
\label{eq:CE}
\begin{aligned}
\frac{1}{N}\sum_{i=1}^N \mathcal{H}({P}_{K \setminus \{g\}}(z_{i}))
\end{aligned}
\end{equation}


\noindent where $N$ is the total number of samples, $K$ is the set of labels.  For the $i^{th}$ sample, $z_{i}$ is the vector of logits for the sample. Let $g$ be the corresponding ground-truth class for the $i^{th}$ sample, so $K \setminus \{g\}$ represents the set of incorrect classes. We use $\mathcal{H}$ to annotate the Shannon entropy function~\cite{doi:10.1002} over the probability $P_{K \setminus \{g\}}(z_{i})$, defined below.

\begin{equation}
\label{eq:h}
    \mathcal{H}(P_{K \setminus \{g\}}(z_{i}))=-\sum_{j} \bigg(P_{K \setminus \{g\}}(z_{i})_{j}\bigg)\log\bigg(P_{K \setminus \{g\}}(z_{i})_{j}\bigg)
\end{equation}
\\
\noindent where $j\in K \setminus \{g\}$, and the probability function $P_{K \setminus \{g\}}(z_{i})_{j}$ is defined as the output of the softmax function:

\begin{equation}
\label{eq:def_p}
    P_{K \setminus \{g\}}(z_{i})_{j}= \frac{e^{z_{i,j}}}{\sum_{k \in K \setminus \{g\}} e^{z_{i,k}}}.
\end{equation}
\\
Intuitively, $P_{K \setminus \{g\}}(z_{i})_{j}$ is the $j^{th}$ dimension of a multinomial distribution normalized among the incorrect classes over logits $z_{i}$ (that is, excluding the probability mass of the ground-truth class). Please note that the alternative definition of complement entropy is mathematically equivalent to the one presented in~\cite{chen2018complement}.   






Despite the good performance by maximizing Complement entropy to make complement events
equally like to occur, this approach do not consider the generalization gap between predicted distributions and true data distributions. For example, if the ground-truth class is ``dog'', to flatten the
predicted probabilities on irrelevant classes such as ``cat'' and ``truck'' is counter-intuitive.








\paragraph{Hierarchical Complement Entropy.} The proposed Hierarchical Complement Entropy (\HCE) regulates the probability masses, similar to what the complement entropy does, but in a hierarchical fashion. Let a subgroup $G$ be a set that contains the sibling classes that belong to the same parental class of the ground-truth class, that is, $g \in G$ and $G \subseteq K$. \HCE will first regulate complement entropy between the subgroup $G$ and the ground truth $g$ followed by the complement entropy between label space $K$ and subgroup $G$. Detailed definition can be found in Eq(\ref{eq:def_p}). The proposed \HCE is defined as the following with $\theta$ being the model parameters:





\begin{equation}
\label{eq:HCE}
\begin{aligned}
HCE(\theta)=\frac{1}{N}\sum_{i=1}^N \big[ \mathcal{H}({P}_{G \setminus \{g\}}(z_{i})) +  \mathcal{H}({P}_{K \setminus G}(z_{i})) \big]
\end{aligned}
\end{equation}
\\
It is not hard to see that Eq(\ref{eq:HCE}) is a direct implementation of the predicted probabilities trained with \HCOT procedure in Figure~\ref{FIG:probability}, which impose probability regulation based on the hierarchical structure of the labels. $\mathcal{H}({P}_{G \setminus \{g\}}(z_{i}))$ regulates inner hierarchy, which corresponds to the relationship between the probability masses marked as red and green. The second term, $\mathcal{H}({P}_{K \setminus G}(z_{i}))$, regulates the outer hierarchy, which corresponds to the relationship between the green and blue class labels. Hierarchical complement entropy ensures that the gaps between each of the hierarchies are as wide as possible to enforce the hierarchical structure during training. In the extreme case when $K=G$, the second term in Eq(\ref{eq:HCE}) disappears and the Hierarchical complement entropy degenerates to Complement entropy.

\paragraph{Optimization.}
Our loss function consists of two terms: the normal cross entropy term (i.e., $XE(\theta)$), and the complement objective term $HCE(\theta)$.

\begin{equation}
\label{eq:loss}
    \mathcal{L} (\theta) = XE(\theta) - HCE(\theta)
\end{equation}
\\
In \textit{Direct optimization}, we simply add two objectives together to be Eq(\ref{eq:loss}) and directly optimize it using SGD without any hyper-parameter tuning or balancing. An alternative approach is \textit{Alternative optimization}, which optimizes the cross-entropy term and the complement objective term interleaved. This is done by maximizing HCE followed by minimizing XE for a single training iteration, which follows \cite{chen2018complement} to have fair comparisons. In our paper, we choose between these two methods to achieve the best performance for our models.

\section{Image Classification}
\label{sec:Experiments_classification}

\begin{figure*}[!t]
\centering
\subfloat[Embeddings trained with cross-entropy]{
    \begin{minipage}[t]{0.48\linewidth}
        \centering
        \includegraphics[width=\linewidth]{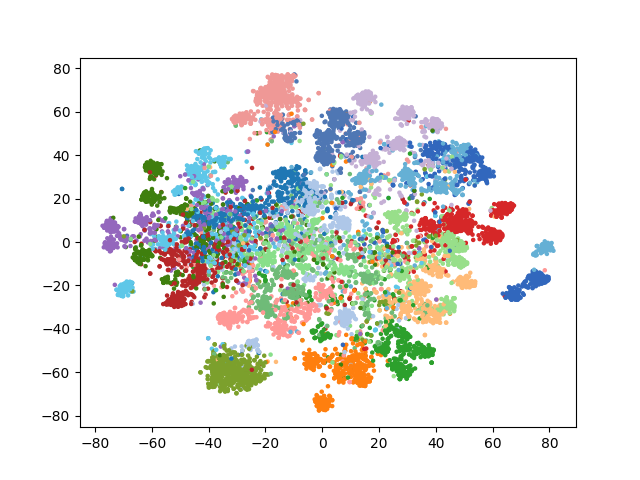}\\
        \vspace{0.02cm}
        \label{fig:tsne_concat_baseline}
    \end{minipage}
    }
\subfloat[Embeddings trained with \HCOT]{
    \begin{minipage}[t]{0.48\linewidth}
        \centering
        \includegraphics[width=\linewidth]{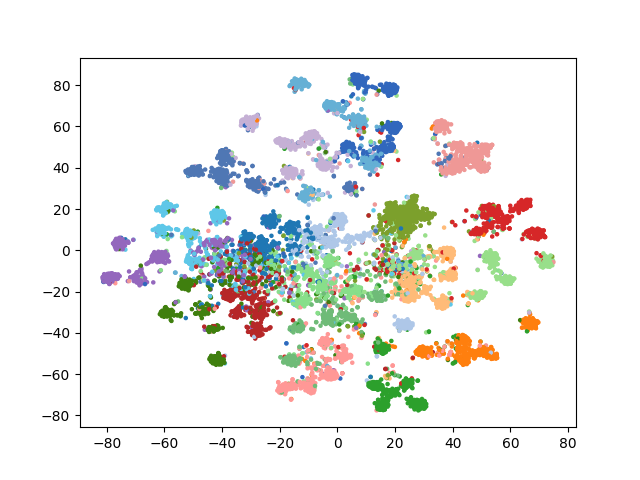}\\
        \vspace{0.02cm}
        \label{fig:tsne_concat_HCOT}
    \end{minipage}
    }

\centering
\caption{Embeddings from 20 coarse classes of CIFAR-100 test images. The embedding of each sample is from the penultimate layer and projected to two dimensions (by t-SNE) for visualization. Notice in (b) the clusters are more distinct, with cleaner and well-separated boundaries, by which we conjecture that the model generalizes better.}
\label{FIG:tsne_concat}
\end{figure*}

In this section, we evaluate HCOT on image classification tasks. Experiments are conducted with two widely-used datasets that contain label hierarchy: CIFAR-100~\cite{Krizhevsky09learningmultiple} and ImageNet-2012~\cite{krizhevsky2012imagenet}. 

We conduct extensive experiments on CIFAR-100 dataset to study several aspects of \HCOT:
\begin{itemize}
    \item Does \HCOT improve the performance over the state-of-the-art models?
    \item Can \HCOT work in synergy with other commonly-used regularization techniques such as Mixup and Cutout?
    \item Does \HCOT improve the classification accuracy of coarse classes over the state-of-the-art models?
    \item How will HCOT affect the latent representation (embedding) of a model? 
\end{itemize}

In addition, we also perform a side-by-side comparison between \HCOT and one state-of-the-art---CNN-RNN \cite{Guo2018}--- that also uses label hierarchy to train image classifiers. The experimental results confirm the proposed \HCOT better captures label structure and learns a more accurate model.

\subsection{CIFAR-100} 
CIFAR-100 is a dataset consisting of 60k colored natural images of 32x32 pixels equally divided into 100 classes. There are 50k images for training and 10k images for testing. The official guide CIFAR-100~\cite{Krizhevsky09learningmultiple} further groups the 100 classes into 20 coarse classes where each coarse class contains five fine classes, forming the label hierarchy. Therefore, each image sample has one fine label and one coarse label. Here we follow the standard data augmentation techniques~\cite{He_2016_CVPR} to preprocess the dataset. During training, zero-padding, random cropping, and horizontal mirroring are applied to the images. For the testing images, we use the original images of $32\times32$ pixels. 


\paragraph{Experimental Setup.} For CIFAR-100, we follow the same settings as the original ResNet paper~\cite{He_2016_CVPR}. Specifically, the models are trained using SGD optimizer with momentum of 0.9; weight decay is set to be 0.0001 and learning rate starts at 0.1, then being divided by 10 at the 100\textsuperscript{th} and 150\textsuperscript{th} epoch. The models are trained for 200 epochs, with a mini-batch size of 128. For training WideResNet, we follow the settings described in~\cite{BMVC2016_87}, and the learning rate is divided by 10 at the 60\textsuperscript{th}, 120\textsuperscript{th} and 180\textsuperscript{th} epoch. In addition, no dropout~\cite{JMLR:v15:srivastava14a} is applied to any baseline according to the best practices in~\cite{batch_norm}. We follow alternating training ~\cite{chen2018complement}, where models are trained by alternating between the primary objective (i.e., cross-entropy) and the complement objective (i.e., Hierarchical Complement Entropy).


\paragraph{Results.} Our method demonstrates improvements over all of the state-of-the-art models compared to baseline and \COT, improving error rates by a significant margin. These models range from the widely used ResNet to the SE-ResNet~\cite{Hu_2018_CVPR}, which is the winner of the ILSVRC 2017 classification competition. SE-ResNet considers novel architecture units named Squeeze-and-Excitation block (SE block) in ResNet framework for explicitly capturing the inter-dependencies between channels of convolutional layers. Results are shown in Table~\ref{CIFAR100 Original ResNet}.

\begin{table}[!ht]
\begin{center}
\begin{tabular}[t]{llll}
\toprule
\multicolumn{1}{l}{ Model} &\multicolumn{1}{l}{Baseline} &\multicolumn{1}{l}{COT}&\multicolumn{1}{l}{\bf \HCOT}
\\\midrule
ResNet-56 ~\cite{He_2016_CVPR}          &29.41 &27.76 &\bf27.3  \\
ResNet-110  ~\cite{He_2016_CVPR}         &27.93 &27.24 &\bf26.46 \\
SE-ResNet-56  ~\cite{Hu_2018_CVPR}         &28.11 &27.04 &\bf26.54  \\
SE-ResNet-110  ~\cite{Hu_2018_CVPR}         &26.49 &26.09 &\bf25.49  \\
PreAct ResNet-18 ~\cite{DBLP:conf/eccv/HeZRS16}    &25.44 &24.73 &\bf23.8  \\
ResNeXt-29 (2$\times$64d) ~\cite{DBLP:conf/cvpr/XieGDTH17}   &23.45 &21.9 &\bf21.64 \\
WideResNet-28-10  ~\cite{BMVC2016_87}   &21.91 &20.99 &\bf20.32 \\
\bottomrule
\end{tabular}
\caption{Error rates (\%) on CIFAR-100 using ResNet, SE-ResNet, and variants of ResNet.}
\label{CIFAR100 Original ResNet}
\end{center}
\end{table}


\paragraph{Results with Mixup and Cutout.} We also show that \HCOT can be applied in synergy with other commonly-used techniques to further improve model performance. We conduct experiments on ResNet-110 with ``Cutout''~\cite{cutout} for input masking and ``Mixup
''~\cite{mixup} for data augmentation.
Table~\ref{CIFAR100 Regularization} shows the accuracy of models trained with \HCOT consistently outperform the baseline and the models
trained with \COT.

\begin{table}[!ht]
\begin{center}
\begin{tabular}[t]{llll}
\toprule
\multicolumn{1}{l}{ Model} &\multicolumn{1}{l}{Baseline} &\multicolumn{1}{l}{COT}&\multicolumn{1}{l}{\bf \HCOT}
\\\midrule
ResNet-110 + Cutout          &24.61 &23.93 &\bf23.85 \\
ResNet-110 + Mixup          &24.46 &23.82 &\bf23.33 \\
\bottomrule
\end{tabular}
\caption{Error rates (\%) on CIFAR-100 using ResNet with Cutout and Mixup techniques.}
\label{CIFAR100 Regularization}
\end{center}
\end{table}

\paragraph{Analysis on Coarse-level Labels.} To understand the places where performance improvements of \HCOT coming from, we show the results by splitting them into coarse and fine labels in Table~\ref{20 coarse labels on CIFAR100}. Here we see that \HCOT improves the performance significantly on the coarse-level labels, where \COT hardly improves. Such a performance improvement is a direct result of modeling label hierarchies, which is not taken into account in either baseline or \COT. Surprisingly, \HCOT also improves fine-level labels significantly, over the already improved results of \COT. This suggests that modeling of the fine-level labels can benefit from modeling label hierarchies. 


\begin{table}[!ht]
\begin{center}
\begin{tabular}[t]{llll}
\toprule
\multicolumn{1}{l}{Label} &\multicolumn{1}{l}{Baseline} &\multicolumn{1}{l}{COT}&\multicolumn{1}{l}{\bf \HCOT}
\\\midrule
Coarse           &15.08 &15.05 &\bf14.02  \\
Fine           &24.21 &23.33 &\bf22.64 \\
\bottomrule
\end{tabular}
\caption{Error rates (\%) on both coarse and fine classes on CIFAR-100 using SE-PreAct ResNet-18.}
\label{20 coarse labels on CIFAR100}
\end{center}
\end{table}

\paragraph{Embedding Space Visualization.} A visualization of logits of the coarse-level labels are shown in Figure~\ref{FIG:tsne_concat}. Here we compare it against the visualisation from the baseline SE-PreAct ResNet-18~\cite{Hu_2018_CVPR} trained using cross-entropy. Compared to the baseline, the \HCOT seems to form more distinct clusters in the embedding space that have clear separable boundaries, by which we conjecture that the model generalizes better and therefore achieves better performance.

\paragraph{Comparison with CNN-RNN.}
To demonstrate the proposed \HCOT effectively leverages label hierarchy, we compare the proposed \HCOT with another state-of-the-art---CNN-RNN \cite{Guo2018}--- that also leverages label hierarchy for training an image classifier. Specifically, CNN-RNN framework is also proposed to take advantage of label hierarchy using an novel neural architecture: combining CNN with RNN. CNN is in charge of extracting discriminative features from images and RNN enables the joint optimization by using coarse and fine labels. In the CNN-RNN framework, WideResNet-28-10 (denoted as WRN) has been selected as the base model and another RNN is constructed upon the WRN. For a fair comparison, we evaluate \HCOT on the same WRN architecture, and the experimental results are provided in Table \ref{CNN-RNN and H-COT on CIFAR100}. The proposed \HCOT achieves significantly better accuracy than WRN-RNN, confirming that the proposed \HCE effectively captures the information from label hierarchy. Also, notice WRN-\HCOT is more parameter efficient---WRN-\HCOT requires less parameters than WRN-RNN since WRN-RNN requires a whole RNN on top of WRN. 


\begin{table}[!ht]
\begin{center}
\begin{tabular}[t]{lll}
\toprule
\multicolumn{1}{l}{Method} &\multicolumn{1}{l}{WRN-RNN} &\multicolumn{1}{l}{WRN-HCOT}
\\\midrule
Top-1 Error          &21.57 &\bf20.32  \\
\bottomrule
\end{tabular}
\caption{Error rates (\%) on CIFAR-100 using ``WRN-RNN'' and ``WRN trained with the proposed HCOT training paradigm'' (denoted as WRN-HCOT).}
\label{CNN-RNN and H-COT on CIFAR100}
\end{center}
\end{table}

\subsection{ImageNet-2012}


ImageNet-2012~\cite{krizhevsky2012imagenet} is a large-scale dataset for image classification with 1k fine categories. 
This dataset consists of approximately 1.3 million training images and 50k validating images, and each image has $256\times256$ pixels. 
In addition, the image labels of ImageNet-2012 are from the ``leaf classes'' of WordNet~\cite{Miller95wordnet:a}; WordNet is a lexical database for the English language, which organizes words into hierarchies defined by hypernym or IS-A relationships.
We follow the prior art on object detection (YOLO9000~\cite{Redmon_2017_CVPR}) to construct hierarchies for labels in ImageNet-2012. Specifically, leaf classes which belong to the same sub-tree are grouped together, and their parental synsets are extracted as the parental classes, forming a two-level hierarchy containing synsets as parental classes and leaf (or fine-grained) classes. 
As a matter of fact, many literature~\cite{Guo2018, 7298619} that aims at improving ImageNet-2012 classification use similar pre-processing steps to construct a tree-based hierarchy into two-level hierarchy.

\paragraph{Experimental Setup.} To prepare for experiments, we apply random crops and horizontal flips during training, while images in the testing set use $224\times224$ center crops (1-crop testing) for data augmentation~\cite{He_2016_CVPR}. We follow~\cite{goyal} as our experimental setup: 256 minibatch size, 90 total training epochs, and 0.1 as the initial learning rate starting that is decayed by dividing 10 at the 30th, 60th and 80th epoch. We use the same alternating training as we did in the CIFAR-100 dataset~\cite{chen2018complement}.  

\paragraph{Results.}
As the main result, we conduct \HCOT with 52 coarse categories. Results in Table~\ref{ImageNet2012 Pure ResNet} shows significant improvements on both top-1 and top-5 error rates compared to COT and the baseline (ResNet-50 using cross-entropy). We note that top-5 error in-explicitly tests the model's abilities for hierarchical labels. 

\begin{table}[!ht]
\begin{center}
\begin{tabular}[t]{llll}
\toprule
\multicolumn{1}{l}{} &\multicolumn{1}{l}{Baseline} &\multicolumn{1}{l}{COT}&\multicolumn{1}{l}{\bf \HCOT}
\\\midrule
Top-1 Error       &24.7 &24.4 &\bf24.0 \\
Top-5 Error      &7.6 &7.4 &\bf7.1 \\
\bottomrule
\end{tabular}
\caption{Validation error rates (\%) on ImageNet-2012 using ResNet-50.}
\label{ImageNet2012 Pure ResNet}
\end{center}
\end{table}

    

\begin{figure*}[!ht]
\centering
\subfloat[Real image]{
    \begin{minipage}[t]{0.23\linewidth}
        \centering
        \includegraphics[scale=0.23]{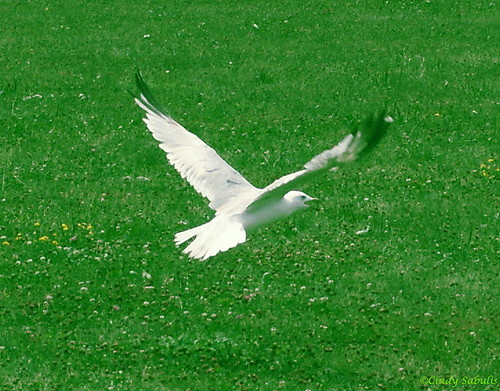}\\
        \vspace{0.05cm}
        \includegraphics[scale=0.23]{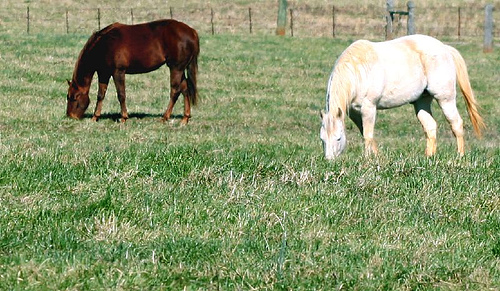}\\
        \vspace{0.05cm}
        \includegraphics[height=1.5in, width=1.6in]{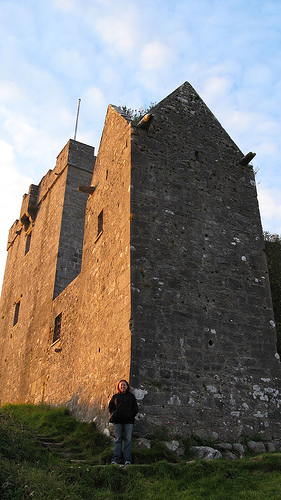}\\
        \vspace{0.05cm}
        \includegraphics[scale=0.23]{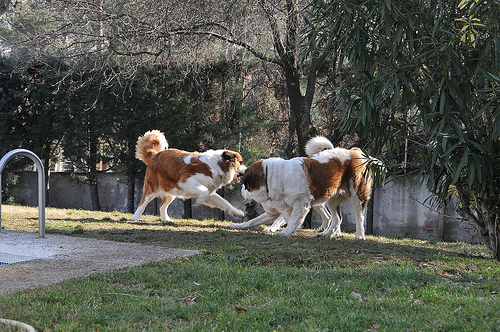}\\
        \vspace{0.05cm}
        \label{fig:compare_fig_a}
    \end{minipage}
    }
\subfloat[Ground truth]{
    \begin{minipage}[t]{0.23\linewidth}
        \centering
        \includegraphics[scale=0.23]{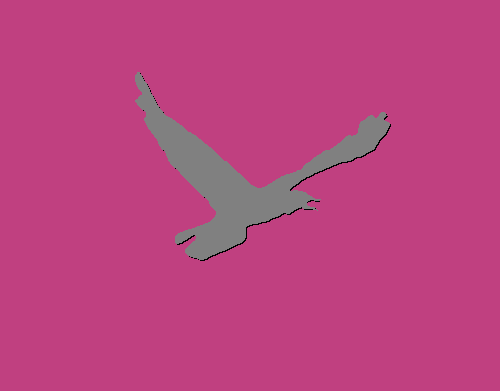}\\
        \vspace{0.05cm}
        \includegraphics[scale=0.23]{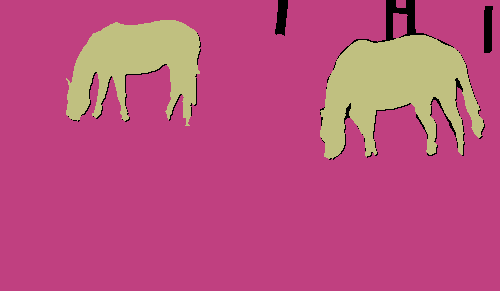}\\
        \vspace{0.05cm}
        \includegraphics[height=1.5in, width=1.6in]{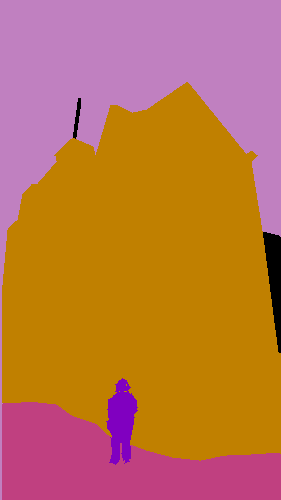}\\
        \vspace{0.05cm}
        \includegraphics[scale=0.23]{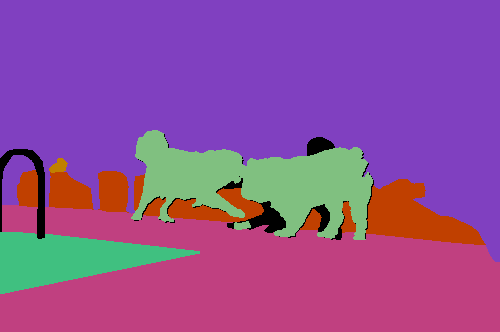}\\
        \vspace{0.05cm}
        \label{fig:compare_fig_b}
    \end{minipage}
    }
\subfloat[EncNet+JPU]{
    \begin{minipage}[t]{0.23\linewidth}
        \centering
        \includegraphics[scale=0.23]{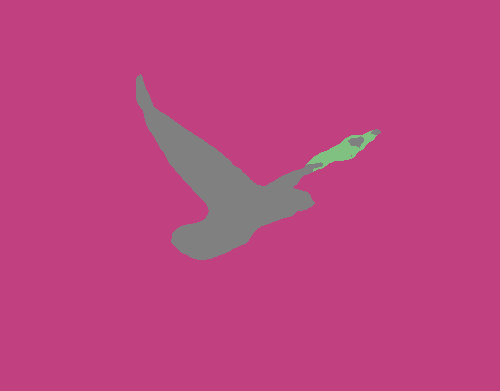}\\
        \vspace{0.05cm}
        \includegraphics[scale=0.23]{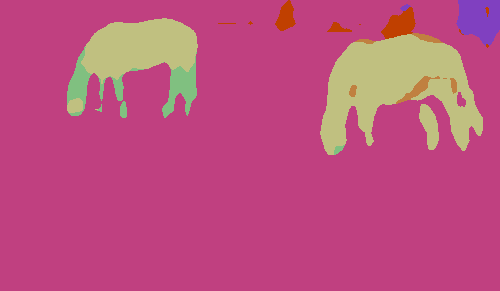}\\
        \vspace{0.05cm}
        \includegraphics[height=1.5in, width=1.6in]{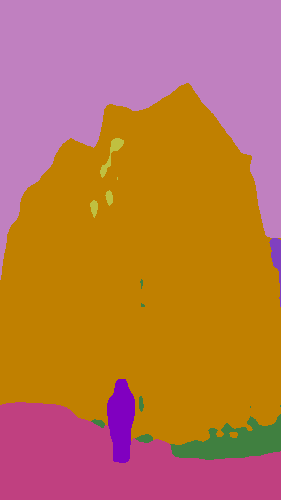}\\
        \vspace{0.05cm}
        \includegraphics[scale=0.23]{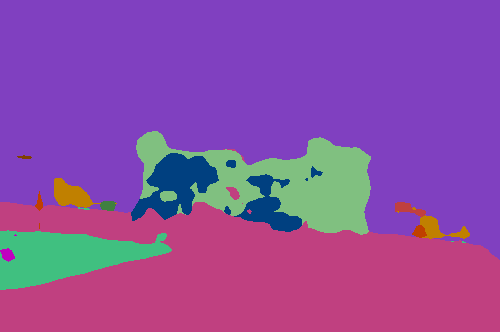}\\
        \vspace{0.05cm}
        \label{fig:compare_fig_c}
    \end{minipage}
}
\subfloat[EncNet+JPU+HCOT]{
    \begin{minipage}[t]{0.23\linewidth}
        \centering
        \includegraphics[scale=0.23]{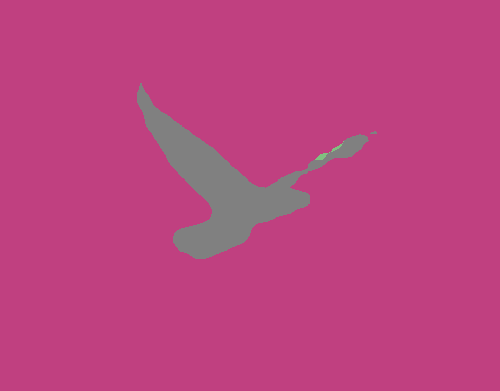}\\
        \vspace{0.05cm}
        \includegraphics[scale=0.23]{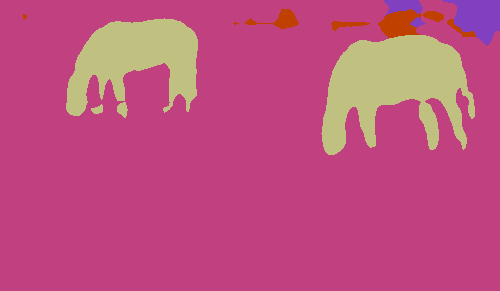}\\
        \vspace{0.05cm}
        \includegraphics[height=1.5in, width=1.6in]{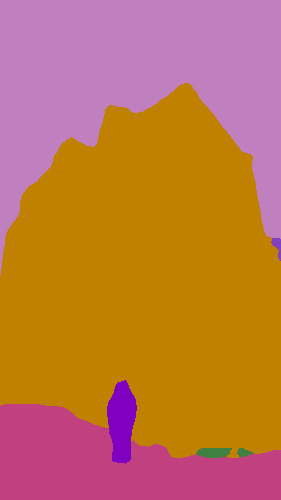}\\
        \vspace{0.05cm}
        \includegraphics[scale=0.23]{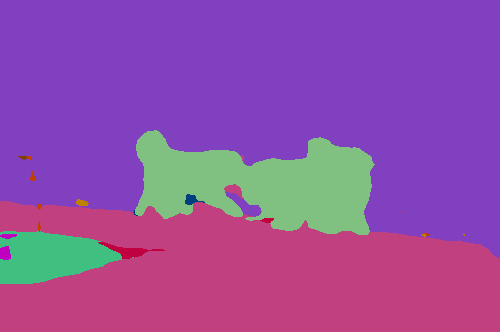}\\
        \vspace{0.05cm}
        \label{fig:compare_fig_d}
    \end{minipage}
}

\centering
\caption{Examples of segmentation results in PASCAL-Context. Notice the main objects (e.g., bird, horses, building, and dogs) in each image are well-segmented in (d) and visually much closer to the ground truth in (b). We believe that \HCOT retains the latent label hierarchies so the segmentation is clearer and without many irrelevant semantics.}
\label{fig:compare_fig}
\end{figure*}

\paragraph{Ablation study.} To explore the effect to \HCOT over different granularity of the coarse classes, we conduct a study on performance of our model over a range of coarse classes $N_{c}$ = \{1, 20, 52, 145, 1000\} on ImageNet-2012. We observe that \HCOT perform best over the sufficient information of category hierarchies. In this case, the performance on Top-1 error peaks at $N_{c}=52$. When the $N_{c}$ goes to the extremes (i.e., 1 or 1000), \HCOT degrades to \COT. This can be illustrated in Eq(\ref{eq:HCE}). When $N_{c}=1$, the left term in the equation will disappear, making it the same as \COT. Similarly, when $N_{c}=1000$, the right term will disappear as there are only 1000 classes.

\begin{table}[!ht]
\begin{center}
\begin{tabular}[t]{llllll}
\toprule
\multicolumn{1}{l}{$N_{c}$} &\multicolumn{1}{l}{1} &\multicolumn{1}{l}{20} &\multicolumn{1}{l}{52}&\multicolumn{1}{l}{145} &\multicolumn{1}{l}{1000}
\\\midrule
Top-1 Error      &24.4 &24.3 &\bf24.0 &24.2 &24.4\\
\bottomrule
\end{tabular}
\caption{Validation error (\%) of different numbers of coarse classes on ImageNet-2012 using ResNet-50.}
\label{Ablation study on the amount of coarse labels conducted on ImageNet-2012}
\end{center}
\end{table}

\section{Semantic Segmentation}
\label{sec:Experiments_segmentation}

In the task of semantic segmentation, there is latent, hierarchical information contained among the labels~\cite{Zhang_2018_CVPR}.
In this task, label hierarchy is not defined or given explicitly, but are rather inferred from the dataset. Applying \HCOT to this task can make effective use of this inferred information in the label space. In particular, the proposed \HCOT procedure can achieve both high confidence of ground-truth and attention of global scene information for each label, which maintains the hierarchy between each semantic and the corresponding theme in a same image sample and helps to provide more accurate semantic segmentation.


\hide{
In standard training process of semantic segmentation, the network is learned from isolated pixels (per-pixel cross-entropy loss for given input image and ground truth labels). Due to the innate hierarchical between the global information from natural images and the local information from pixels in the tasks of semantic segmentation, we conjecture applying \HCOT on these tasks can simply make the model learn hierarchical information from loss function and try to control the complement distribution for avoiding competitions at the boundary and make the segmentation more accurate. Therefore, we apply \HCOT on the widely-used Pascal-Context dataset~\cite{6909514} and demonstrate the significant improvements.
}

\hide{
For applying \HCOT for image segmentation, we investigated the stability and the performance of performing experiments on two training procedures: ``alternative training procedure'' and ``added term''. The results show that simply using hierarchical complement objective as added term in existing SOTA segmentation loss function can stably achieve significant improvements.  
}

\begin{table*}[!ht]
\label{Segmentation results}
\centering
\begin{minipage}{\textwidth}
    \subfloat[EncNet]{
    \begin{minipage}[h]{0.49\textwidth}
    \centering
    \begin{tabular}[t]{lll}
    \toprule
    \multicolumn{1}{l}{Method} &\multicolumn{1}{l}{PixAcc} &\multicolumn{1}{l}{mIoU$\%$}
    \\\midrule
    XE           &0.7835 &49.70  \\
    \COT          &0.7844 &49.65   \\
    \HCOT        &\bf0.7862 &\bf49.86 \\
    \bottomrule
    \label{Segmentation results using EncNet on Pascal Context dataset}
    \end{tabular}
    \end{minipage}
    }
    \subfloat[EncNet+JPU]{
    \begin{minipage}[h]{0.49\textwidth}
    \centering
    \begin{tabular}[t]{lll}
    \toprule
    \multicolumn{1}{l}{Method} &\multicolumn{1}{l}{PixAcc} &\multicolumn{1}{l}{MIoU$\%$}
    \\\midrule
    XE           &0.7880 &51.05  \\
    \COT          &0.7884 &51.07 \\
    \HCOT           &\bf0.7918 &\bf51.35 \\
    \bottomrule
    \label{Segmentation results using Joint Pyramid Upsampling based on EncNet on Pascal Context dataset}
    \end{tabular}
    \end{minipage}
    }
\end{minipage}
\caption{Segmentation results of models trained with cross-entropy (denoted as XE) versus \COT and \HCOT on PASCAL-Context dataset.}
\end{table*}


\paragraph{Dataset.}
We apply \HCOT on the widely-used ``Pascal-Context" dataset~\cite{6909514}. Each image in the PASCAL-Context dataset has dense, semantic labels over the entire scene of the image. The dataset contains 4,998 images for training and 5,105 for testing. We follow the prior arts~\cite{ChenPK0Y16, LinMS016, 6909514} and create a set of 60 semantic labels for segmentation.  These 60 semantic labels represent the most frequent 59 object categories, plus the ``background'' category.

\paragraph{Experimental Setup.}
We first take EncNet (Context Encoding Module)~\cite{Zhang_2018_CVPR} to be the baseline. Here we follow the previous works~\cite{DBLP:journals/corr/ChenPSA17, DBLP:journals/corr/YuKF17, Zhao_2017_CVPR} to use the dilated network strategy on the pretrained ResNet-50. In addition, we perform the Joint Pyramid Upsampling (JPU)~\cite{abs-1903-11816} instead of the dilated convolution over EncNet (denoted as ``EncNet+JPU'') to reproduce the state-of-the-art results in semantic segmentation. JPU can formulate the task of extracting high resolution feature maps into a joint upsampling problem. 8The training details are the same as described in~\cite{abs-1903-11816, Zhang_2018_CVPR}.  We train the model for 80\textsuperscript{th} epochs with SGD and set the initial learning rate as 0.001. The images are then cropped to $480\times480$ and grouped with batch size 16. For data augmentation, we randomly left-right flip and scale the image between 0.5 to 2. We also use the polynomial learning rate scheduling as mentioned in~\cite{Zhao_2017_CVPR}. Different from the training procedure on classification, here we adopt \textit{direct optimization} which training our model by combing the complement loss and primary loss together, which achieves a better empirical performance and only needs marginal extra computation cost compared to baselines.



\paragraph{Evaluation.} We use the pixel accuracy (PixAcc) and mean Intersection of Union (mIoU) as the evaluation metrics with single scale evaluation. Specifically, for the PASCAL-Context dataset, we follow the procedure in the standard competition benchmark~\cite{zhou2017scene} and calculate mIoU by ignoring the pixels that are labeled as ``background''.


\hide{We perform EncNet~\cite{Zhang_2018_CVPR}, which called Context Encoding Module as the strong baseline and also conduct Joint Pyramid Upsampling (JPU)~\cite{abs-1903-11816} on EncNet for comparing to the state-of-the-art results in semantic segmentation. For training details, we mostly follow the protocol from~\cite{Zhang_2018_CVPR,abs-1903-11816}. Specifically, 
we use Resnet-50, which is widely used in most existing segmentation methods as our backbone model. We train the model for 80\textsuperscript{th} epochs with SGD and set the initial learning rate as 0.001. The images are then cropped to $480 × 480$ and grouped with batch size 16. We also follow the prior work~\cite{Zhao_2017_CVPR} to use the polynomial learning rate scheduling. 
}





\paragraph{Results.}
We evaluate the quality of the segmentation from the models trained with pixel-wise cross-entropy (as baseline) and trained with \HCOT, by quantitatively calculating the PixAcc and mIoU scores and visually inspecting the output image segments. Specifically, to make sure the improvement on the segmentation from \HCOT comes from leveraging the label hierarchies, we have also conducted the experiment on the models trained with \COT to perform a three-way comparison. Experimental results show that \HCOT achieves better performance than \COT and baseline (cross-entropy) as shown in Table~\ref{Segmentation results using EncNet on Pascal Context dataset}. We also form ``EncNet+JPU'' as another baseline, and the \HCOT again significantly outperforms \COT and cross-entropy (as shown in Table~\ref{Segmentation results using Joint Pyramid Upsampling based on EncNet on Pascal Context dataset}). As segmentation does not have inherent label hierarchies, hierarchical structures among labels will have to be inferred from the data. Images occur frequently together as a theme will in-explicitly form a label hierarchy that will be learned to improve the performance of the model. 

\paragraph{Visualizations.}
In Figure~\ref{fig:compare_fig}, we show segmentation results from three test images on PASCAL-Context dataset. In addition to the input images (Figure~\ref{fig:compare_fig_a}), we show the ground-truth segmentation (Figure~\ref{fig:compare_fig_b}) and the results from EncNet+JPU model trained with cross-entropy (Figure~\ref{fig:compare_fig_c}) and trained with the proposed \HCOT (Figure~\ref{fig:compare_fig_d}). The segments generated by the proposed \HCOT are less fragmented and have less noises.

\section{Conclusion}
\label{sec:Conclusion}
In this paper, we propose Hierarchical Complement Objective Training (HCOT) to answer the motivational question. HCOT is a new training paradigm that deploys Hierarchical Complement Entropy as the training objective to leverage information from label hierarchy. \HCOT neutralizes the probabilities of incorrect classes at different granularity: under the same parental category as the ground-truth class or not belong to the same branch. HCOT has been extensively evaluated on image classification and semantic segmentation tasks, and experimental results confirm that models trained with HCOT significantly outperform the state-of-the-arts. A straight-line future work is to extend HCOT into other computer vision tasks which involve rich information of latent label hierarchies but still unexplored.

\clearpage

{\small
\bibliographystyle{ieee_fullname}
\bibliography{egbib}
}

\end{document}